\documentclass{article}

\usepackage{arxiv}

\usepackage[utf8]{inputenc}
\usepackage{url}
\usepackage{comment}

\usepackage{longtable}

\usepackage{graphicx}
\graphicspath{ {./drawings/} }

\newcommand{\todo}[1]{}
\newcommand{\figcite}[1]{{#1}}

\begin{document}

\title{Modular Design Patterns\\for Hybrid Learning and Reasoning Systems:\\
a taxonomy, patterns and use cases}

\author{Michael van Bekkum[1] \And Maaike de Boer[1] \And Frank van Harmelen\thanks{Frank.van.Harmelen@vu.nl}~~[2] \And Andr\'{e} Meyer-Vitali[1] \And Annette ten Teije\thanks{Annette.ten.Teije@vu.nl}~~[2]\\[2ex]
~[1] TNO, The Netherlands\\
~[2] Vrije Universiteit Amsterdam, The Netherlands 
}

\maketitle

\begin{abstract}
The unification of statistical (data-driven) and symbolic (knowledge-driven) methods is widely recognised as one of the key challenges of modern AI. 
Recent years have seen large number of publications on such hybrid neuro-symbolic AI systems. That rapidly growing literature is highly diverse and mostly empirical, and is lacking a unifying view of the large variety of these hybrid systems. In this paper we analyse a large body of recent literature and we propose a set of \emph{modular design patterns} for such hybrid, neuro-symbolic systems. We are able to describe the architecture of a very large number of hybrid systems by composing only a small set of elementary patterns as building blocks.
The main contributions of this paper are: 1) a taxonomically organised vocabulary to describe both processes and data structures used in hybrid systems; 2) a set of 15+ design patterns for hybrid AI systems, organised in a set of elementary patterns and a set of compositional patterns; 3) an application of these design patterns in two realistic use-cases for hybrid AI systems.  Our patterns reveal similarities between systems that were not recognised until now. Finally, our design patterns extend and refine Kautz' earlier attempt at categorising neuro-symbolic architectures. 
\end{abstract}


\section{Introduction}
\label{sec:motivation}
It is widely acknowledged in recent AI literature that the data-driven and knowledge-driven approaches to AI have complementary strengths and weaknesses \cite{Darwiche:CACM2018}. This has led to an explosion of publications that propose different architectures to combine both symbolic and statistical techniques. Surveys exist on narrow families of such systems \cite{von2019informed,Wang:2017,asim2018survey}, but to date no conceptual framework is available in which such hybrid symbolic-statistical systems can be discussed, compared, configured and combined.
In this paper, we propose a set of modular design patterns for Hybrid AI systems that combine learning and reasoning (using combinations of data-driven and knowledge-driven AI components). With this set of design patterns, we aim to achieve the following goals. First, we provide high-level descriptions of the architectures of hybrid AI systems. Such abstract descriptions should enable us to better understand the commonalities and differences between different systems, while abstracting from their specific technical details. We will show for a number of our design patterns that they describe systems which have not been recognised in the literature as essentially doing the same task. Secondly, our set of design patterns is intended to bridge the gap between the different communities that are currently studying hybrid approaches to AI systems. AI communities such as machine learning and knowledge-based systems as well as other communities (such as cognitive science) often use very different terminologies which hamper the communication about the systems under study. Finally, and perhaps most importantly, our design patterns are modular, and are intended as a tool for engineering hybrid AI systems out of reusable components. In this respect, our design patterns for hybrid AI systems have the same goals as the design patterns which are well known from Software Engineering \cite{DesignPatterns:1994}.

Darwiche \cite{Darwiche:CACM2018} draws attention to the distinction between two types of components in AI systems, for which he uses the terms function-based and model-based. Similarly, Pearl \cite{PearlImpediments:2018} uses the term "model-free" for the representations typically used in many learning systems. Other names used for inferences at this layer are: ``model-blind'', ``black-box'', or ``data-centric'' \cite{PearlImpediments:2018}, to emphasize that the main task performed by machine learning systems is function-fitting: fitting data by a complex function defined by a neural network architecture. Such "function-based" or "model-free" representations are in contrast to the "model-based" representations typically used in reasoning systems. We will use a similar distinction in our design patterns. Both Darwiche and Pearl argue for combining components of these two types: "the question is not whether it is functions or models but how to profoundly integrate and fuse functions with models'' \cite{Darwiche:CACM2018} and  "Our general conclusion is that human-level AI cannot emerge solely from model-blind learning machines; it requires the symbiotic collaboration of data and models" \cite{PearlImpediments:2018}. However, neither of the cited works discuss how such combinations must be made. This is exactly what we set out to do in this paper by proposing a set of modular design patterns for such combinations.

Lamb \& Garcez \cite{garcez2020neurosymbolic} ask for \textit{``The trained network and the logic [to] become communicating modules of a hybrid system [..]. This distinction between having neural and symbolic modules that communicate in various ways and having translations from one representation to the other in a more integrative approach to reasoning and learning should be at the centre of the debate in the next decade.''}
Our proposal is a concrete step in precisely the direction that they call for, providing a set of composition patterns that can be used as modular building blocks in hybrid systems.

The main contributions of this paper are as follows: (i) a taxonomically organised vocabulary to describe both the processes that constitute hybrid AI systems as well as the data structures that such processes produce and consume; (ii) a set of modular design patterns for hybrid AI systems, organised in a set of elementary patterns, plus more sophisticated patterns that can be constructed by composing such elementary patterns; we will show how a number of systems from the recent literature can be described as such compositional patterns; these patterns are a further elaboration of our first proposal for such design patterns in \cite{VanHarmelen2019Boxology}; (iii) two realistic use-cases for hybrid AI systems (one for skill matching, one for robot action selection), showing how the architecture for each of these use-cases can be described in terms of our compositional design patterns.

The paper is structured as follows, first we describe our taxonomical vocabulary in section \ref{sec:taxonomy}, in section  \ref{sec:elementary-patterns} we describe a set of elementary patterns, followed by a set of compositional patterns in section \ref{sec:compositional-patterns}. Section \ref{sec:use-cases} describes two realistic use-cases in terms of the patterns from section \ref{sec:elementary-patterns} and \ref{sec:compositional-patterns}. Section \ref{sec:future-work} concludes and discusses directions for future work.

\section{A Taxonomical Vocabulary}
\label{sec:taxonomy}
In order to describe design patterns, a terminology is required that defines a taxonomy of both processes and their inputs and outputs on various levels of abstraction. On the highest level of abstraction, we define instances, models, processes and actors. In the pattern diagrams, instances are represented as rectangular boxes, models as hexagonal boxes, processes as ovals and actors as triangles.
More specific concepts will be used, when necessary and useful, using a colon-separated notation. For example, \textit{model:stat:NN} refers to a neural network model. The level of abstraction depends on the use of the pattern and the stage of design and implementation: the closer to implementation, the more specific the concepts will be. The design patterns should abstract from implementation details, but be specific enough to document applicable design choices. As abbreviated notation, we will not name the highest abstraction level, because that is implied by the type of box. In the models, we always indicate whether a specific model is statistical (\textit{stat}) or semantic (\textit{sem}).
The full taxonomy with definitions can be found in the Appendix.

\subsection{A Taxonomy of Instances}

Instances are the basic building blocks of ``things'', examples or single occurrences of something. The two main classes of instances are data and symbols. The precise distinction between symbols and ``non-symbols'' remains a contentious issue in modern philosophy. We are following \cite{Berkeley:2008} by imposing the following requirements before any token can be deemed a symbol: (1) a symbol must designate an object, a class or a relation in the world, and such a designated object, class or relation is then called the interpretation of the symbol; (2) symbols can be either atomic or complex, in which case they are composed of other symbols according to a formal set of compositional rules; and (3) there must be system of operations that, when applied to a symbol, generates new symbols, that again must have a designation. Thus, the tokens p1 and p2 may designate particular persons, with the symbol r designating some relation between these persons; then r(p1,p2) is a complex symbol made out of these atomic symbols, and the operation r(p1,p2) $\models_T$ r(p2,p1) defines an operation constructing one complex symbol out of another. In logic, such ``operations'' correspond to logical inference $\models$, and this logical view is most relevant in this paper, but in another context such operations may be transitions between the symbols that denote the states of a finite state machine. All this makes the symbol p1 different from a data item, say a picture of a person, where the collection of pixels may be an accurate image of a person, but does not designate the person in a way that allows the construction of more complex designations and operations that transform these into other designations. Simply put: a symbol p designates a person, whereas a picture is just itself. Such tokens which are not symbols are what we will call ``data'' in this paper. 

The types of data that appear in Hybrid AI systems include

\begin{itemize}
\item numbers: numerical measurements;
\item texts: sequences of words, sentences;
\item tensors: multi-dimensional spaces, including bitmaps;
\item streams: (real-time) sequences of data, including video and other sensor inputs.
\end{itemize}
The types of symbols, on the other hand, include
\begin{itemize}
\item labels: short descriptions;
\item relations: connections between data items, such as triples and other n-ary relations;
\item traces: (historical) records of data and events, such as proof traces for explanations.
\end{itemize}

\subsection{A Taxonomy of Models}
Models are descriptions of entities and their relationships. They are useful for inferring data and knowledge. Models in Hybrid AI systems can be either (1) statistical or (2) semantic models.
Statistical models represent dependencies between statistical variables. Examples are (Deep) Neural Networks, Bayesian Networks and Markov Models.
Semantic models represent the implicit meaning of symbols by specifying their concepts, attributes and relationships. Examples are Taxonomies, Ontologies, Knowledge Graphs, Rulebases and Differential Equations. We summarise these semantic models under the umbrella term ``knowledge base'' (KB).

\subsection{A Taxonomy of Processes}

In order to perform operations on instances and models, processes define the steps that lead from inputs to results. Three main types of processes are: (i) the generation of instances and models, (ii) their transformation and (iii) inferencing thereupon.
Generation of models is performed either via training or, ``manually'' by knowledge engineering with experts. Many forms of transformations exist, such as transforming a knowledge graph to a vector space. Inferences are made using induction or deduction. Induction is constructing a generalisation out of specific instances. Such a generalisation (a ``model'') can take many different forms, ranging from the trained weights in a neural network to clauses in a learned Logic Program, but in every case, such models are created by inductions on instances.

Using such models, we can apply deductive inferencing in order to reach conclusions about specific  instances of data.  Commonly, deduction is associated with logical inference, but the standard definition, namely as inference in which the conclusion is of no greater generality than the premises, equally applies to the forward pass of a neural network, where a general model (the trained network) is applied to an instance in order to arrive at a conclusion. Both inductive and deductive processes can be either symbolic or statistical, but in every case induction reasons from the specific (the instances) to the general (a model), and deduction does the converse. Thus, the distinction between induction and deduction corresponds precisely to the distinction between learning and reasoning. Both classification (``the systematic arrangement in groups or categories according to established criteria'', see our appendix) and prediction (``to calculate some future event or condition as a result of analysis of available data'', again see our appendix) are deductive processes, since both use a generic model (obtained through an inductive process of learning or engineering) to derive information about specific instances (the objects to be classified or the events to be predicted).

\subsection{A Taxonomy of Actors}

Processes in an AI system are initiated by autonomous actors, based on their intentions and goals. They interact with each other using many protocols and behaviours, such as collaboration, negotiation or competition. These interactions lead to collective intelligence and emergent social behaviour, such as in swarms, multi-agent systems or human-agent teams. Autonomy is a gradual property, ranging from remotely controlled to selfish behaviour and all forms of cooperation in between. Actors can be humans, (software) agents or robots (physically embedded agents). Examples are proactive software components, apps, services, mobile agents, drones or (parts of) autonomous vehicles.
Actors are not yet used explicitly in the current collection of patterns, but will be included in the future, when we will also dive into distributed AI and human-agent interaction.

\section{Elementary Patterns}
\label{sec:elementary-patterns}

The taxonomy of instances, models and processes from the previous section gives rise to a number of elementary building blocks for hybrid systems that combine learning and reasoning. The ``train'' process consumes either data or symbols to produce a model (all these terms taken from the taxonomy):

\hfill \break
\centerline{\includegraphics[scale=0.6]{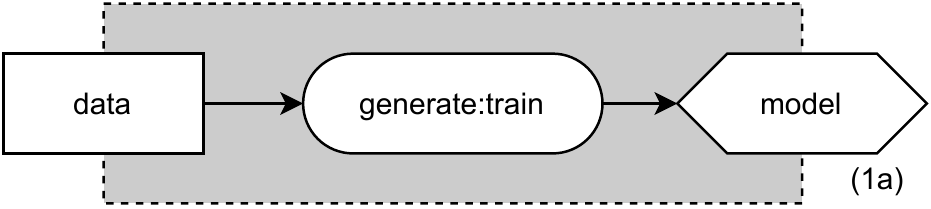}}

\hfill \break
\centerline{\includegraphics[scale=0.6]{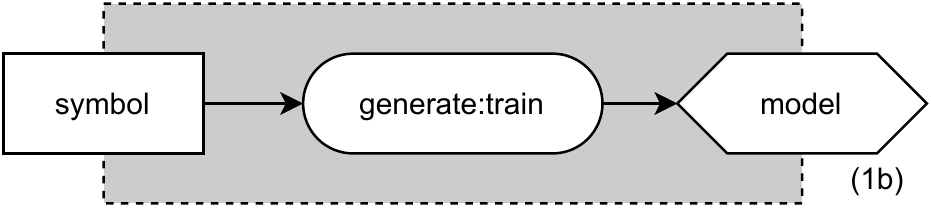}}

Additionally, an actor (e.g., a domain expert or knowledge engineer) can create a model, such as an ontology or rule-base:

\hfill \break
\centerline{\includegraphics[scale=0.6]{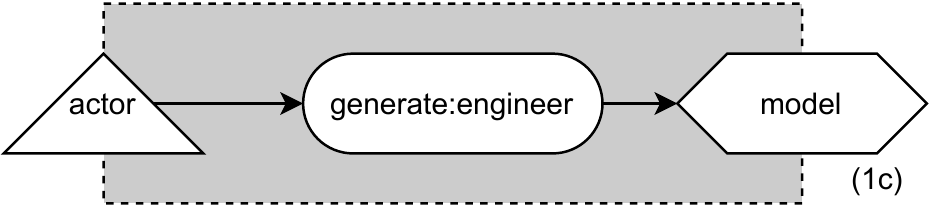}}

In the processing, often a transformation step is needed to create the right type of data, either from symbol or data:

\hfill \break
\centerline{\includegraphics[scale=0.6]{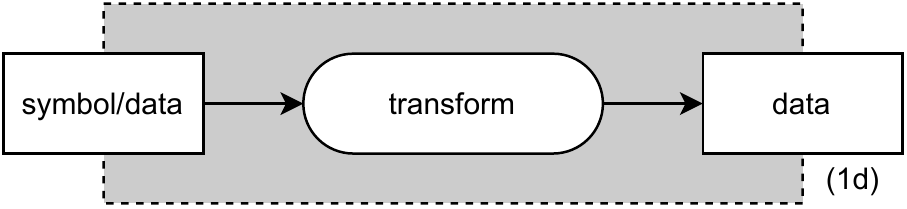}}

Of course models are only trained in order to be subsequently used in a downstream task (predicting labels, deriving conclusions, predicting links, etc). This process is captured in the patterns \figcite{(2a)-(2c)}, depending on the symbolic or statistical nature of the data. Following our taxonomy, an infer step uses such models in combination with either data or symbols to infer conclusions:

\hfill \break
\centerline{\includegraphics[scale=0.6]{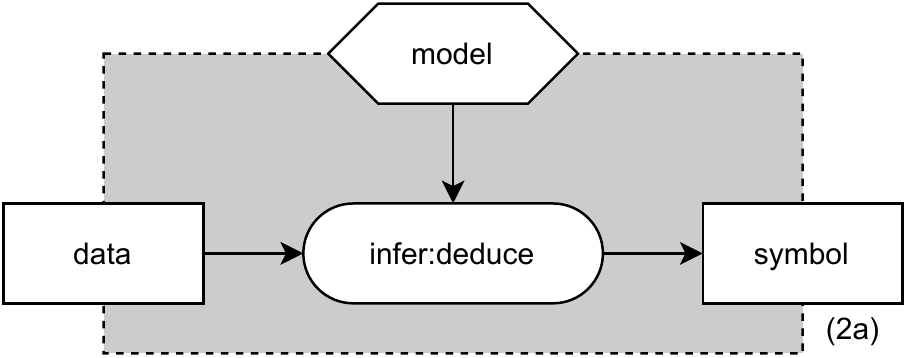}}

\hfill \break
\centerline{\includegraphics[scale=0.6]{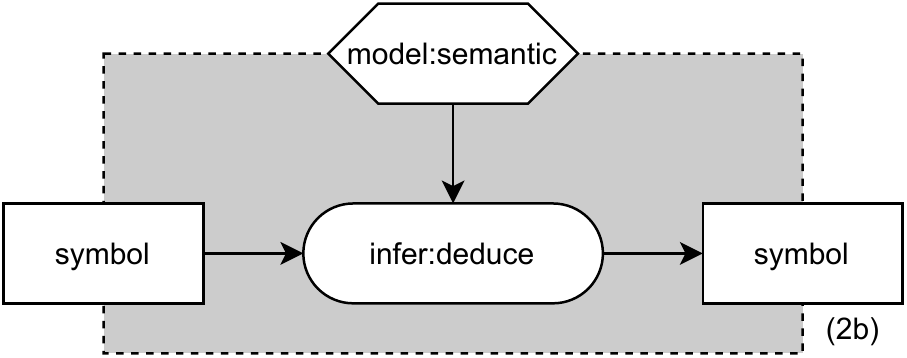}}

\hfill \break
\centerline{\includegraphics[scale=0.6]{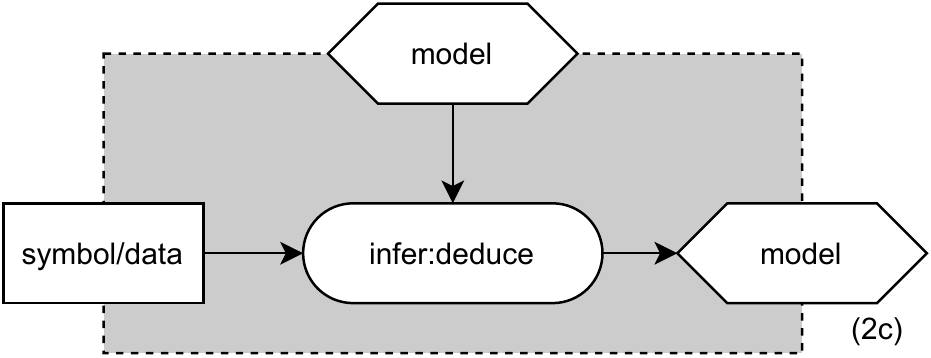}}

Finally, sometimes an operation on a semantic model is neither a logical induction or deduction, but a transformation into another datastructure. This is captured by the final elementary pattern \figcite{(2d)}:

\hfill \break
\centerline{\includegraphics[scale=0.6]{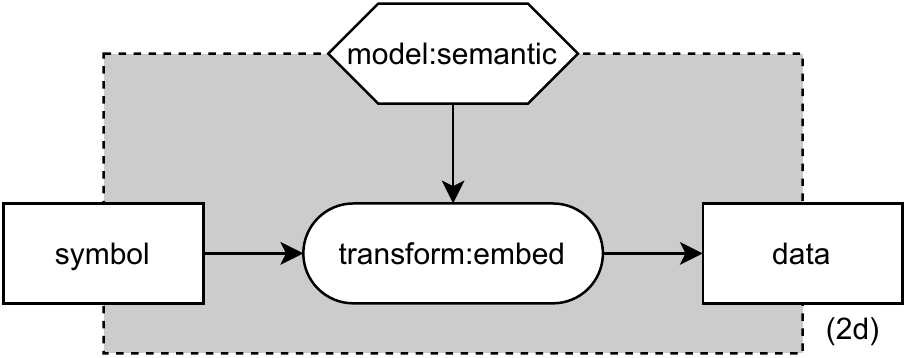}}

As is encoded in these diagrams, the types of models involved (symbolic or statistical) and the type of results derived is constrained by the type of the inputs of these elementary processes. 

These elementary patterns allow us to give a more precise definition of the concept of "hybrid systems", which is often used rather nebulously in the literature:

\textbf{Definition:}
\textit{Machine Learning systems} are systems that combine pattern \figcite{(1a)} with \figcite{(2a)}, yielding pattern \figcite{(3a)}, see below;
\textit{Knowledge Representation systems} are systems that follow pattern \figcite{(2b)};
\textit{Hybrid systems} are systems that form any other combination of the elementary patterns \figcite{(1a)-(2d)}.

Already these elementary patterns (1a-1d,2a-2d), even in their almost trivial simplicity, can be used to group together a large number of very different approaches from the literature: even though the algorithms and representations of Inductive Logic Programming (ILP) \cite{inoue2017special}, Markov Logic Networks \cite{richardson2006markov} and Probabilistic Soft Logic \cite{PSL:2013,PSL:2017}  are completely different, the architecture pattern \figcite{1b} applies to all of them, showing that they are all aimed at the same goal: learning over symbolic structures. 

Similarly, learning a symbolic rule-set that captures rules for knowledge graph completion \cite{Meilicke:2020} is captured by this pattern. Constructing knowledge graph embeddings into a high-dimensional vector space \cite{Paulheim:JWS2017,Wang:2017,Nickel:2016} is also captured by figure 1b.
So ILP and KG embedding would each be captured more specifically by adding type annotations to the constructed model: \textit{model:sem} for ILP, and \textit{model:stat} for KG embedding.
Many "classical" learning algorithms such as decision tree learning and rule mining, as well as deep learning systems, are  covered by architectural pattern \figcite{1a}. 
The learning patterns \figcite{(1a)-(1b)} must be combined with the prediction patterns \figcite{(2a)-(2c)} to give a model for the full learning and prediction task \figcite{(3a)}: 

% 3

\hfill \break
\centerline{\includegraphics[scale=0.6]{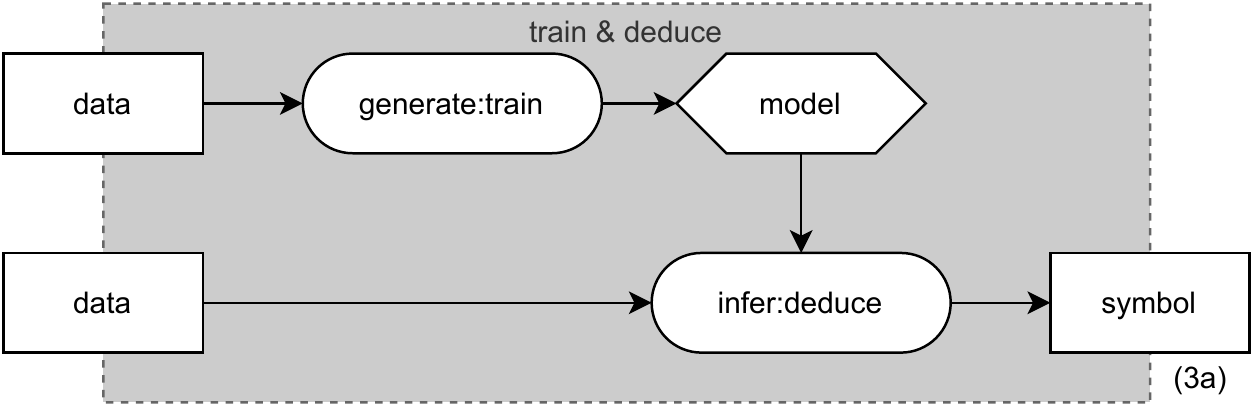}}

\todo{Many of our pictures would become more readable if in all our patterns we would treat (3a) and (3b) as single boxes, instead of being composed of 1a and 2a,b. This would require redrawing the bounding box + internal label in many of our figures.}

This model is precisely the composition of the elementary processes for train and infer given above. Even learning with a regular neural network is captured by this diagram (although it is not typically recognized that the feed-forward phase, when a trained neural network is applied to new data, is actually a deductive task, namely reasoning from given premises (input data plus the learned network weights) to a conclusion. 

Analogously to learning from \emph{data}, it is also possible to learn from \emph{symbols}, using elementary patterns \figcite{(1b)} and \figcite{(2b)} instead of  \figcite{(1a)} and \figcite{(2a)}:

\hfill \break
\centerline{\includegraphics[scale=0.6]{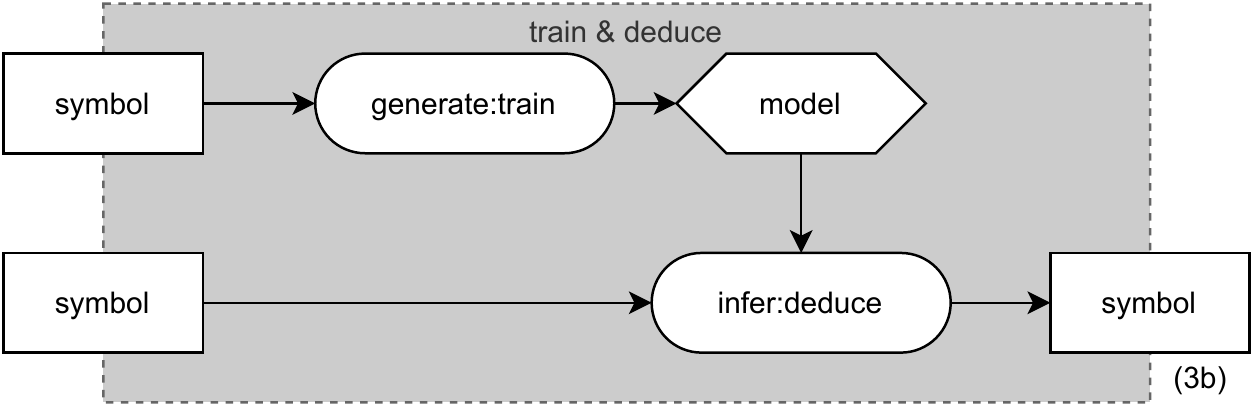}}

%FvH \hfill \break
%FvH \centerline{\includegraphics[scale=0.6]{drawings/Boxology-03c.pdf}}
%FvH \todo{Need to describe (3c) in the text.}

As mentioned above, this pattern then describes the learning (and subsequent inference) in Inductive Logic Programming, Knowledge Graph embeddings, Probabilistic Soft Logic and Markov Logic Networks. 

Using our hierarchical taxonomy, the two patterns \figcite{(3a) and (3b)} can be abstracted into a single pattern, replacing all the boxes labelled with "data" or "symbol" by the generic term "instance". The specific diagrams \figcite{(3a) and (3b)} can then be recovered by adding type annotations "instance:sym" or "instance:data". 
Such type-specialisations maintain the insight that many of these very different approaches (ILP, MLN, PSL, Knowledge Graph embeddings) actually follow the same schema. 
\todo{Patterns (3a,b) are re-used very often, and they complicate the resulting diagrams of which they are part quite a lot. Should we just merge 1a+2a in a single diagram? (either by only dropping the separate bounding boxes for 1a and 2a (and refer to 3a instead), or by merging 1a+2a in a simpler diagram (and refer to 3a) Use fig. 3a-x = 3a.3 with separate 1a/2a boxes }. 

\section{A collection of compositional patterns}
\label{sec:compositional-patterns}

In this section, we describe compositional patterns based on the elementary pattern described in the previous section. We combined papers from several fields into one pattern.
From the elementary patterns, we create compositions in two ways: (1) we can create a more complex pattern by connecting or `stitching' elementary patterns; (2) we can go more specific or more abstract (only showing the boxes of 1a - 1b for example); in a specific pattern we can specify the type of, for example, a symbol block in terms of \textit{symbol:relations}.

\subsection{Learning from data with symbolic output}

\todo{Add "Human-Driven FOL Explanations of Deep Learning" (IJCAI2020) as another example of "learning symbols from data"?}

\hfill \break
\centerline{\includegraphics[scale=0.6]{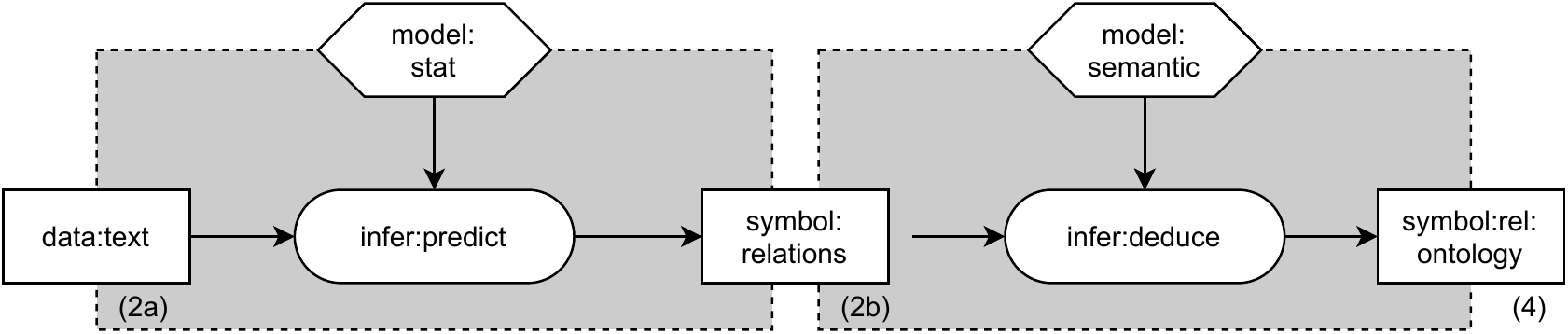}}

In ontology learning, a symbolic ontology is learned from data in the form of text \cite{OntologLearning,asim2018survey,konstantopoulos2010formulating,bouraoui2017inductive,wong2012ontology,brewster2008mind,textminingOntologyLearning}. The text is first translated into (subject, verb, object) relations using a statistical model such as the Stanford Parser \cite{cimiano2009ontology,boer2019creating}. These relations are an intermediate representation. A semantic model, for example rules for Hearst patterns, can then infer the relations that form a full ontology including relation hierarchies and axioms. This pattern combines patterns (2a) and  (2b).  Whereas in other cases an ontology can play the role of a model on the basis of which properties of instances are deduced, in this case, we represent the ontology as a set of relations, because it is the output of a process, and not a model which is input to a process.

A related but different instantiation of this pattern is the use of text-mining not to learn full-blown ontologies, but to learn just the class/instance distinction (which is always problematic in ontology modelling), as done in
\cite{Patel-Schneider:EKAW2018}. As concerns the architectural patterns, this work only differs in the actual content of the symbolic output: a full-blown ontology, or only a class/instance label. 

In contrast, other ontology-learning systems \cite{learningconceptualspaces,konstantopoulos2010formulating} start from a given set of relations (the "A-box" of description logic) and then infer an ontological hierarchy. These systems only apply the second half of the above pipeline, pattern \figcite{(2b)}. 

An entirely different application domain is found in \cite{Asai:2019}, where symbolic first-order logic representations are generated to describe the content of images.

\subsection{Explainable learning systems through rational reconstruction}

%5
\centerline{\includegraphics[scale=0.6]{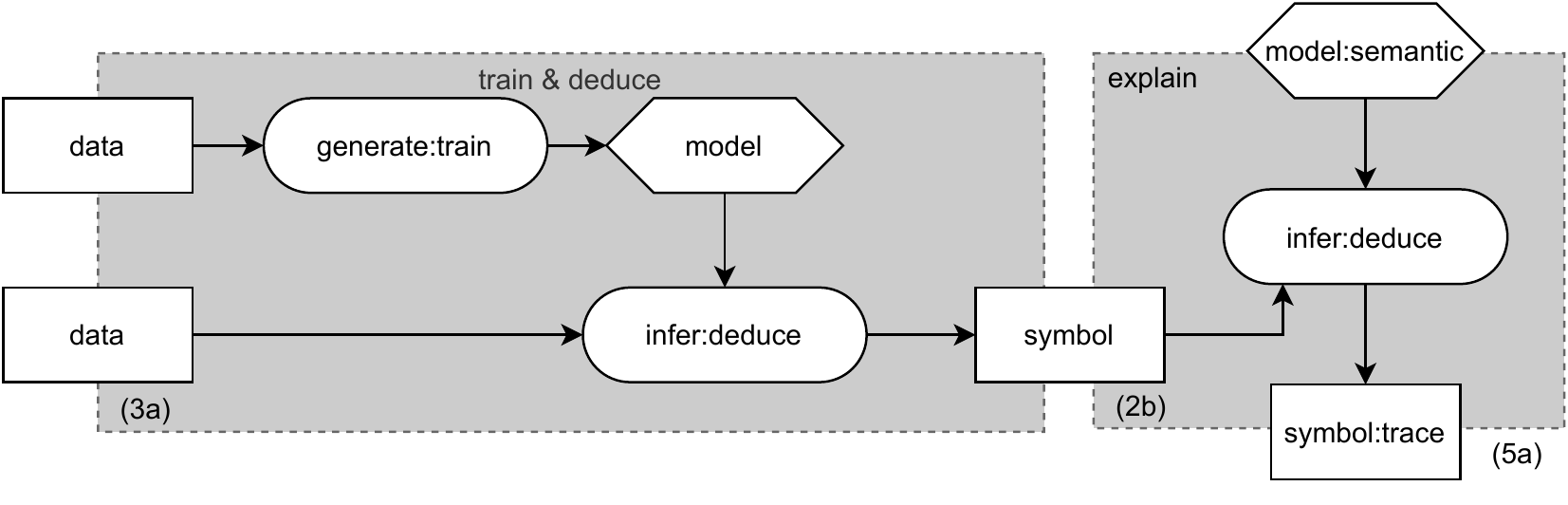}}
\hfill \break
\centerline{\includegraphics[scale=0.6]{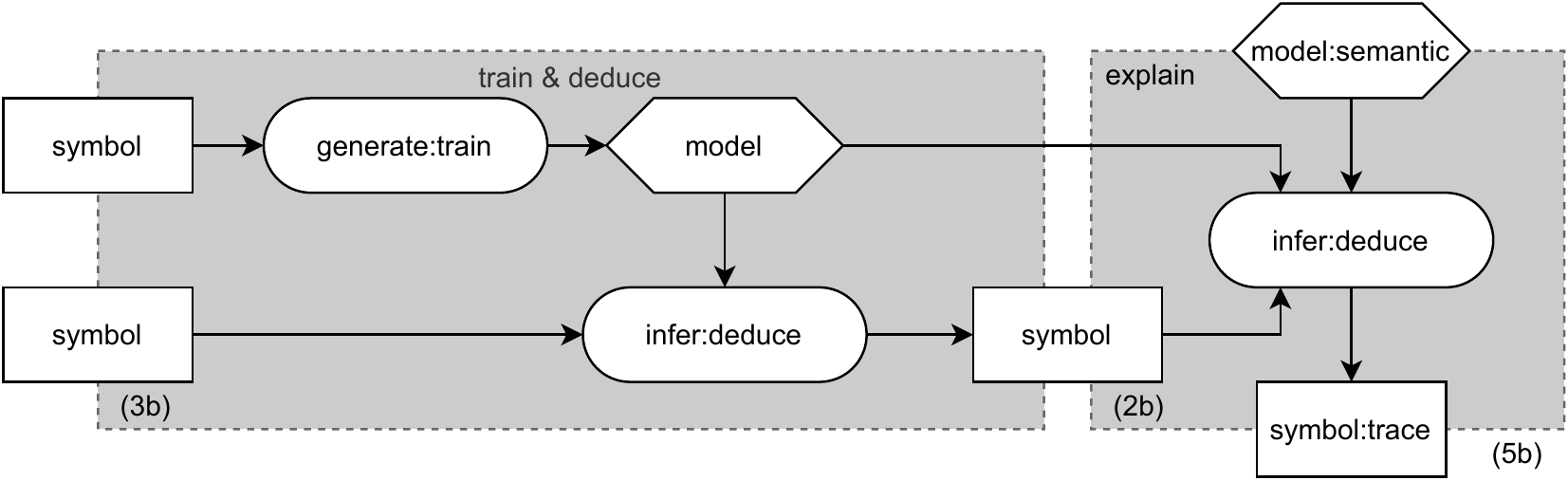}}

Hybrid symbolic-statistical systems are seen as one possible way to remedy the ``black-box'' problem of many modern machine learning systems \cite{WeldIntelligibleAI}. Pattern \figcite{(5a)} shows one of the hybrid architectures that have been proposed in the literature for this purpose. A standard machine learning system is trained {(generate:train)} to construct a model which is then applied to input data in order to produce {(infer:deduce)} a prediction (for example, a label for a given input image). The result of this process (in the form of the pairs of image + label) is then passed on to a symbolic reasoning system which then uses background knowledge {(model:semantic)} to produce a ``rational reconstruction'' of a reason to justify the input/output-pair of the learning system. An example of this is the work by \cite{Tiddi-Explaining} who uses large knowledge graphs to reconstruct the justification of temporal patterns learned from Google Trends data. It is important to emphasize that the justification found by the symbolic system is unrelated to the inner workings of the black box of the machine learning system. The symbolic system produces a post-hoc justification that is not necessarily reflecting the statistical computation. This architecture is also used in \cite{Hitzler-Explaining:2017}, where a description logic reasoner is used to come up with a logical justification of classifications produced by a deep learning system. Notice that pattern (5a) is a straightforward combination of elementary patterns (1a), (2a) and (2b). 

Pattern \figcite{(5a)} captures so-called "instance-level explanations", where a separate explanation is generated for every specific result of applying the learned model. In contrast, it is also possible to generate "model-level explanations", where a generic explanation is constructed that captures the structure of the entire learned model. An example of this is \cite{Ciravegna:2020}, which trains a second neural network that uses the input/output behaviour of a classifier network to general first-order logic formulas that can then be used to explain the behaviour of the classifier. This results in a modification of the above pattern in which the subsystem labelled with "2b" is replaced by a learning system that takes the learned model from 1a as input and produces a symbolic explanation of that model. In other words: the explanation is generated based on the trained model, and not just on the derived individual result of applying that model to a given piece of data.

\todo{Are we missing a 2nd family of explainable systems, namely those based on looking \emph{into} the black box?}

\subsection{Learning an intermediate abstraction}

\centerline{\includegraphics[scale=0.6]{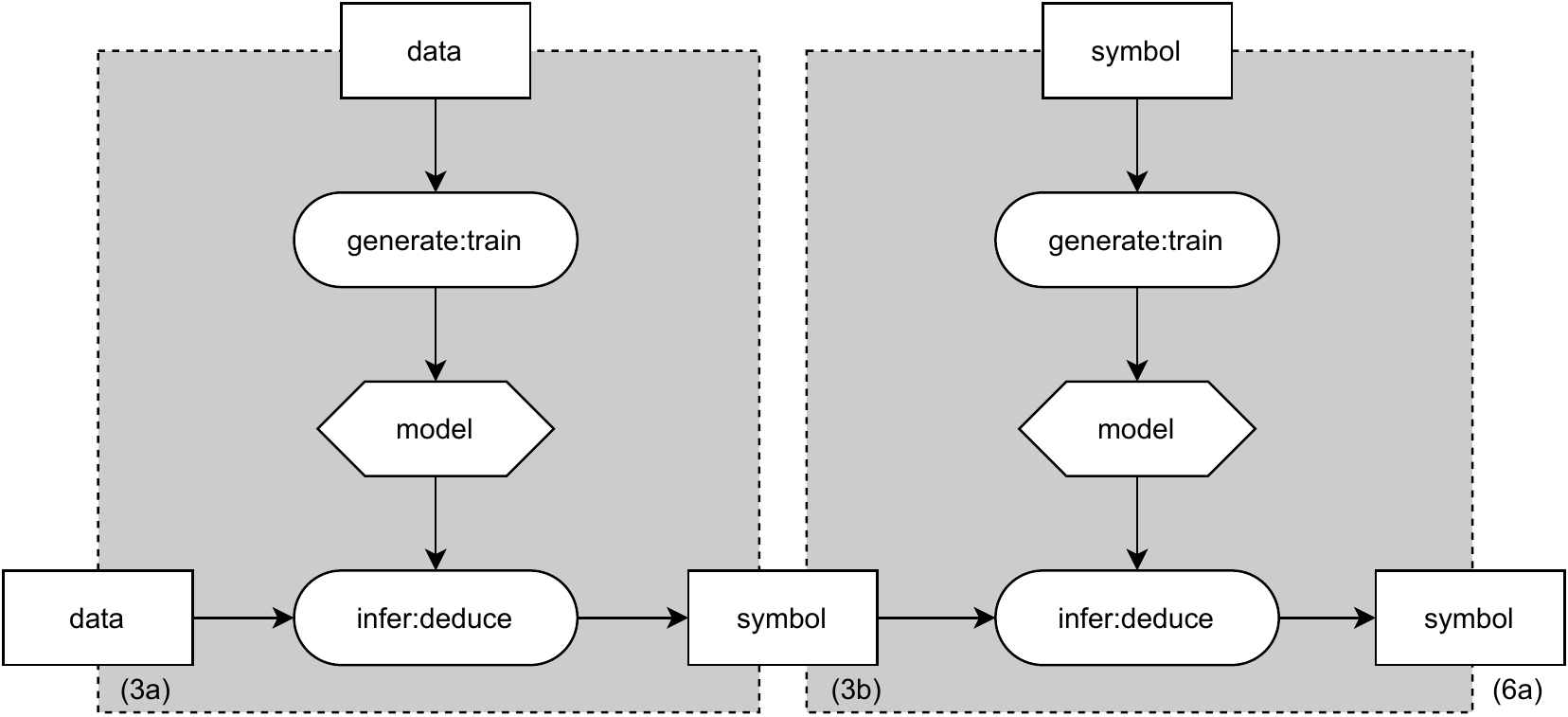}}

\textbf{Intermediate abstraction for learning.}
A well known machine learning benchmark is to recognise sets of handwritten digits \cite{Lecun98}. This digit-recognition task could be extended to perform addition on such handwritten digits through end-to-end training on the bitmap representations of the handwritten digits. This would correspond to our basic pattern (3a). However, many authors have observed that it is more efficient to learn an intermediate symbolic representation (mapping the pixel representation of the handwritten digit into a symbolic representation), and then train a model to solve the symbolically represented addition task. This pattern is represented above, where two standard train+deduce patterns (pattern 3a, 3b) are chained together through a symbolic intermediate representation which serves as output for the first pattern and as input for the second. This pattern is exploited in Deep Problog \cite{DeepProbLog:2018} where a system is trained to add handwritten digits by first recognising the digit and then doing the addition. This then turns out to be a much more robust approach then simple end-to-end training going from the digit bitmaps to the summed result in one step. This same pattern also captures the DeepMind experiment \cite{DeepSymbolicReinforcementLearning:2016} where a reinforcement learning agent is trained to not just navigate on a bitmap representation of its world, but to first learn an intermediate symbolic representation of the world and then use that for navigation. 

Besides learning a spatial abstraction (as in \cite{DeepSymbolicReinforcementLearning:2016}), the work in \cite{MultipleRLtasks} uses the same architecture pattern for deriving a temporal abstraction of  sequence of subtasks, which are then input to reinforcement learning agents. 
One of the advantages of such an intermediate representation is the much higher rates of transfer learning that can be obtained after making trivial changes to the input distribution, be they handwritten digits or bitmaps of floor spaces. 

\textbf{Intermediate abstraction for reasoning}. Whereas pattern (6a) consists of a composition of two patterns for learning, pattern (3) (first deriving an intermediate abstraction on the basis of a trained model and then using this intermediate abstraction as the basis for a further derivation on the basis of a second trained model), it is also possible to use the derived abstraction as the input for a deductive reasoning task, composing pattern (3) with pattern (2b), creating pattern (6b). A classic example of this pattern is the AlphaGo system \cite{AlphaGo:2017}, where machine learning is used to train an evaluation function that gives rise to a symbolic search tree which is then traversed using deductive techniques (Monte Carlo tree search) in order to derive a (symbolically represented) next move on the Go board. {Notice that pattern 5a is a specialisation of this pattern 6b.} 

\centerline{\includegraphics[scale=0.6]{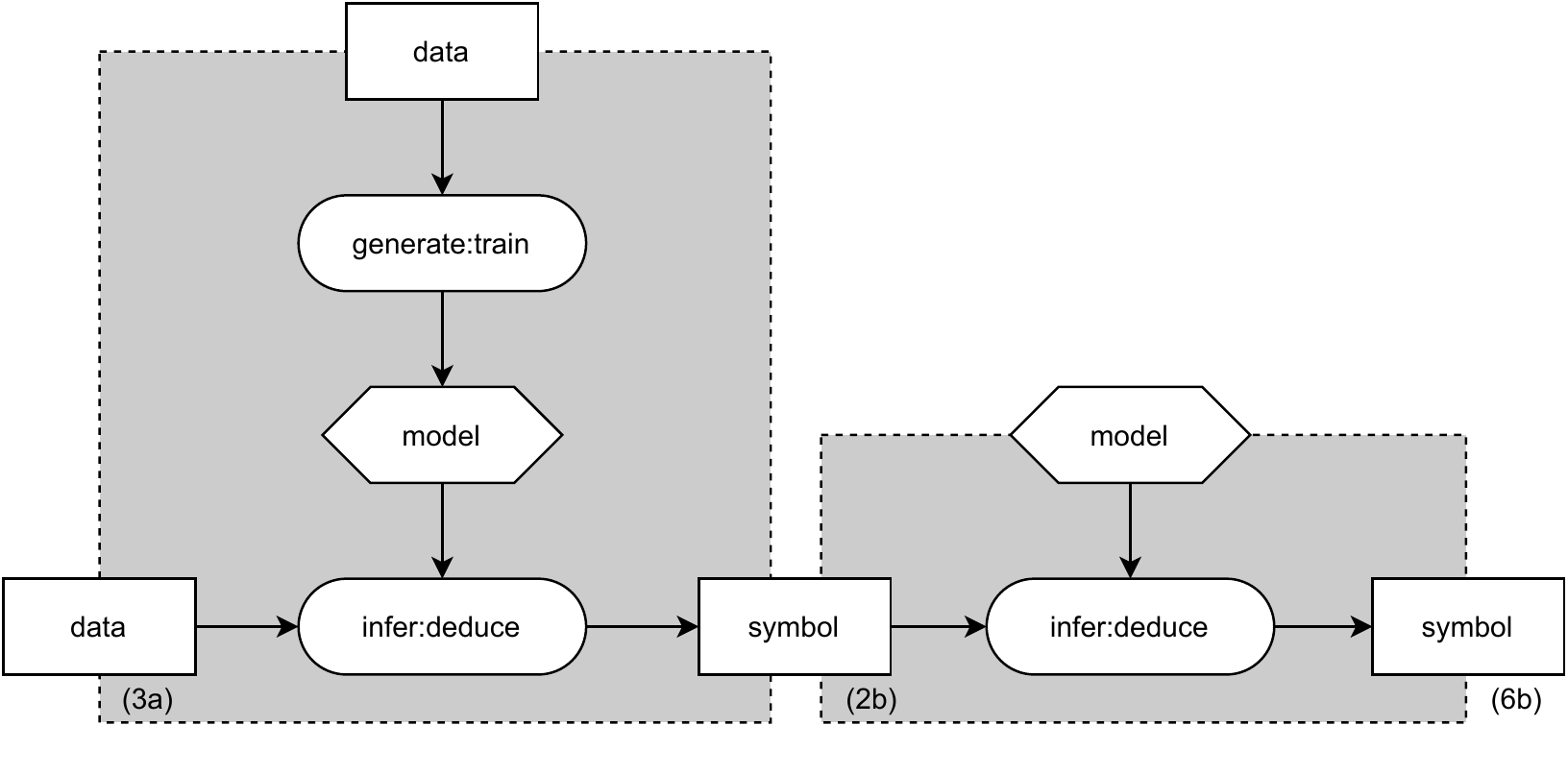}}

Pattern (4) above (ontology learning from text) can now be seen to also be a variation on the general theme of ``learning an intermediate abstraction'', where the set of relations extracted by linguistic analysis is the intermediate abstract that is input for the set of rules that constructs the final ontology out of these relations. In pattern (4) the models {(model:semantic)} are assumed to be given (e.g. by using the pretrained Stanford parser) and hence the training phases are omitted. 

\subsection{Informed learning with prior knowledge}

\centerline{\includegraphics[scale=0.6]{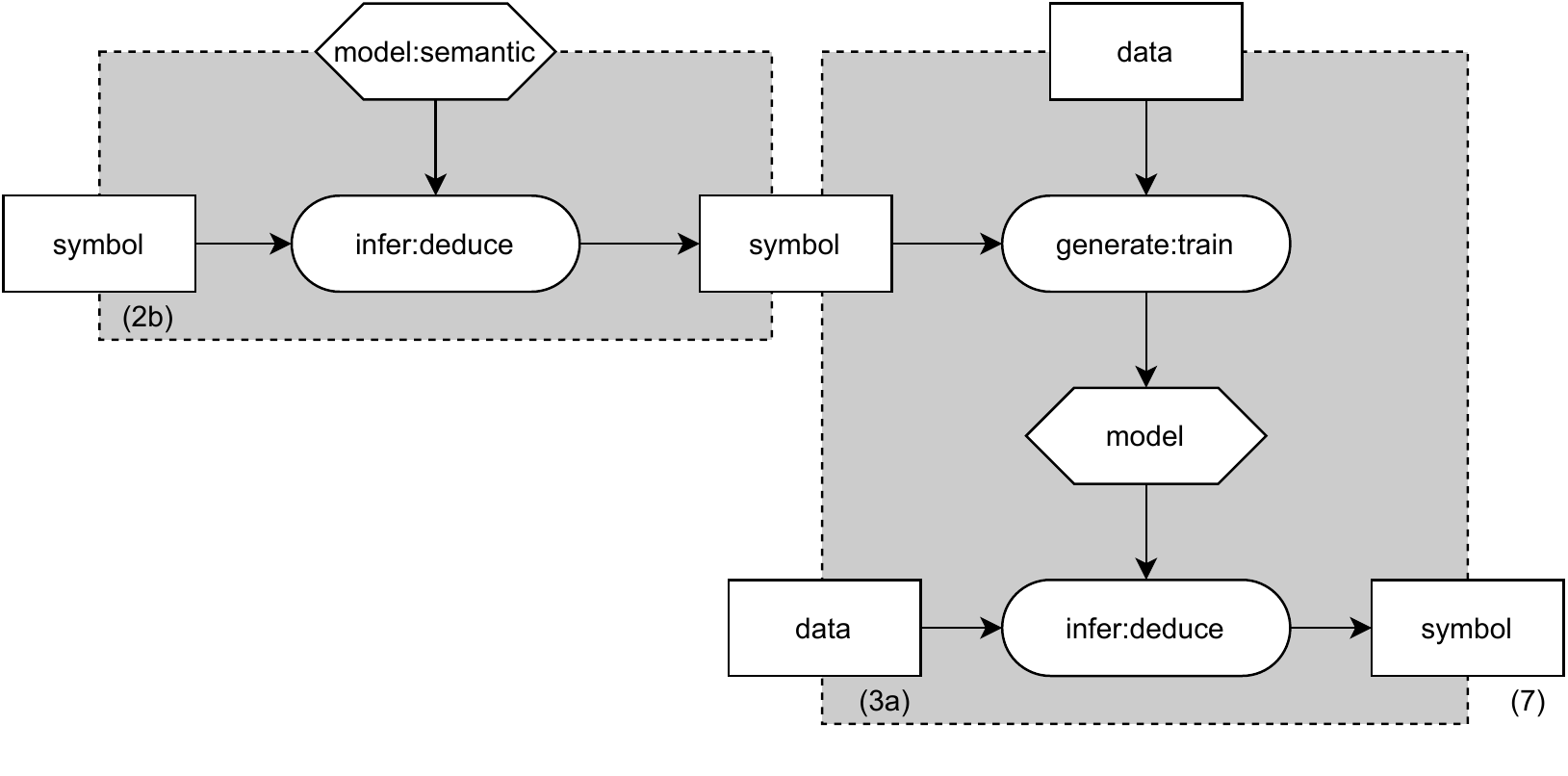}}

In \cite{von2019informed} a large collection of over 100 different systems is discussed which are all captured by pattern \figcite{(7)}. In this pattern, the training phase of a learning system is guided by  information that is obtained from a symbolic inference system (pattern 2b). For this purpose, the training step from elementary pattern \figcite{(1a)} is extended with a further input to allow for this guidance by inferred symbolic information. A particular example is where domain knowledge (such as captured in modern large knowledge graphs) is used as a symbolic prior to constrain the search space of the training phase \cite{Bian:2014}. In general, this patterns also captures all systems with a so-called semantic loss function \cite{semantic-loss-function}, where (part of) the loss-function is formulated in terms of the degree to which the symbolic background knowledge is violated. 
Such a semantic loss-function is also used in \cite{DiLeillo:2020}, where the semantic loss is calculated by weighted model counting. In \cite{Detassis:2020} and \cite{Diligenti:2020}  the semantic loss-function is realised through approximate constraint satisfaction. Another example is \cite{injecting-rules-into-embeddings} where logical rules are used as background knowledge for a gradient descent learning task in a high-dimensional real-valued vector space. In the same spirit, \cite{DeSa:2018} exploits a type-hierarchy to inform an embedding in hyperbolic space. 

Logic Tensor Networks \cite{LTN:IJCAI2017} also fall in this category, since they jointly minimise both the loss function of a neural network and maximise the degree to which a first-order logic theory is satisfied. The fact those LTN's are captured by the same design pattern as semantic loss functions suggests an analogy between the two (namely that the maximisation of first-order satisfiability in LTN's can be regarded as a semantic loss-function). This analogy between these two systems was not mentioned in their original papers, but only comes to light through our analysis in terms of high-level design patterns.

An entirely different category of systems that is captured by the same pattern are constrained reinforcement learners (e.g. \cite{Geibel06}), where the exploration behaviour of a reinforcement learning agent is constrained through symbolic constraints that enforce safety conditions.
Similarly, \cite{Illanes:2020} uses high-level symbolic plans to guide a reinforcement learner towards efficiently learning a policy. \cite{Silvertri:2020} shows how adding domain knowledge in the form of symbolic constraints greatly improves the sampling-frequency of a neural network trained to solve a combinatorial problem. The LYRICS system \cite{LYRICS:2019} proposes a generic interface layer that allows to define arbitrary first order logic background knowledge, allowing a learning system to learn its weights under the constraints imposed by the prior knowledge.

The full design pattern \figcite{(7)} requires that the symbolic prior is derived by a symbolic reasoning system, but it is of course also possible that this symbolic prior (or "inductive bias", using the terminology from \cite{DeepMind:2018}) is simply in the form of an explicit knowledge-base for which no further derivation is possible. This would lead to a simplified version of pattern (7) where the ``Infer'' step would be omitted. An example of this is \cite{Hoogendoorn:2016}, where input data is first abstracted with the help of a symbolic ontology, and is then fed into a classifier, which performs better on the abstracted symbolic data than on the original raw data. A similar example is given in \cite{Tresp:ISWC2017}, where knowledge graphs are successfully used as priors in a scene description task.

An interesting variation is presented in \cite{Zhang:2019}. This work exploits the fact that pattern (7) uses prior knowledge in symbolic as input for learning, while pattern (4) produces symbolic results as the output of learning. The system described in \cite{Zhang:2019} then iterates between producing symbolic knowledge, then using this symbolic knowledge as input for informed machine learning, followed again by using the learned model to produce better symbolic knowledge, hence iterating between patterns (7) and (4). The IterefinE system \cite{IterefinE:2020} is another example of this pattern. 

\subsection{From symbols to data and back again}

\centerline{\includegraphics[scale=0.6]{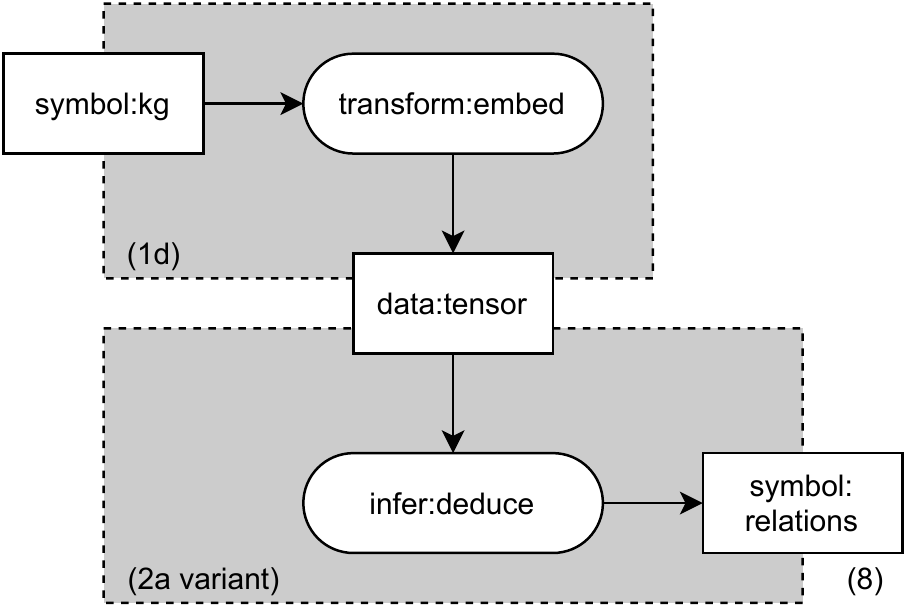}}

Link prediction (or: graph completion) in knowledge graphs \cite{Paulheim:JWS2017,Wang:2017,Nickel:2016}, has been a very active area on the boundary of symbolic and statistical representations, and is an example of what is captured in pattern (8). Almost all graph completion algorithms perform this task by first translating the knowledge graph to a representation in a high-dimensional vector space (a process called ``embedding'', this is captured in pattern (1d)),  and then use  this representation to predict additional edges which are deemed to be true based on geometric regularities in the vector space, even though they are missing from the original graph. {This can be expressed in a variant of pattern 2a.}

\subsection{Learning logical representations for statistical inferencing}

Integrating knowledge representations into a machine learning system is a long standing challenge in Hybrid AI, since it allows logic calculus to be carried out by a neural network in an efficient and robust manner. Encoding prior knowledge also allows for better training results on fewer data. This pattern describes the integration of logic into a machine learning model through tensorization of the logic involved \cite{garcez2019neural}
by applying prior (semantic) knowledge representations as constraints for machine learning. The pattern transforms a semantic model into vector/tensor representations and uses these to train a neural network in order to learn. The machine learner can make inferences based on its embedded logic, for example Logic Tensor Networks \cite{serafini2016logic,zhou2018graph} 
where relational data is embedded in a (convolutional) neural network. Graph Neural Networks (GNNs, \cite{borgwardt2020graph}) embed a semantic graph model by transforming a neighbourhood structure of its nodes and edges into vectors and using these to train a neural network. In \cite{cohen2020scalable} a reified knowledge base is provided as input to train neural modules to perform multi-hop inferencing tasks. The pattern in itself is an extension of pattern 3a, where the training input data is a (transformed) representation of relational data.
\\

\centerline{\includegraphics[scale=0.6]{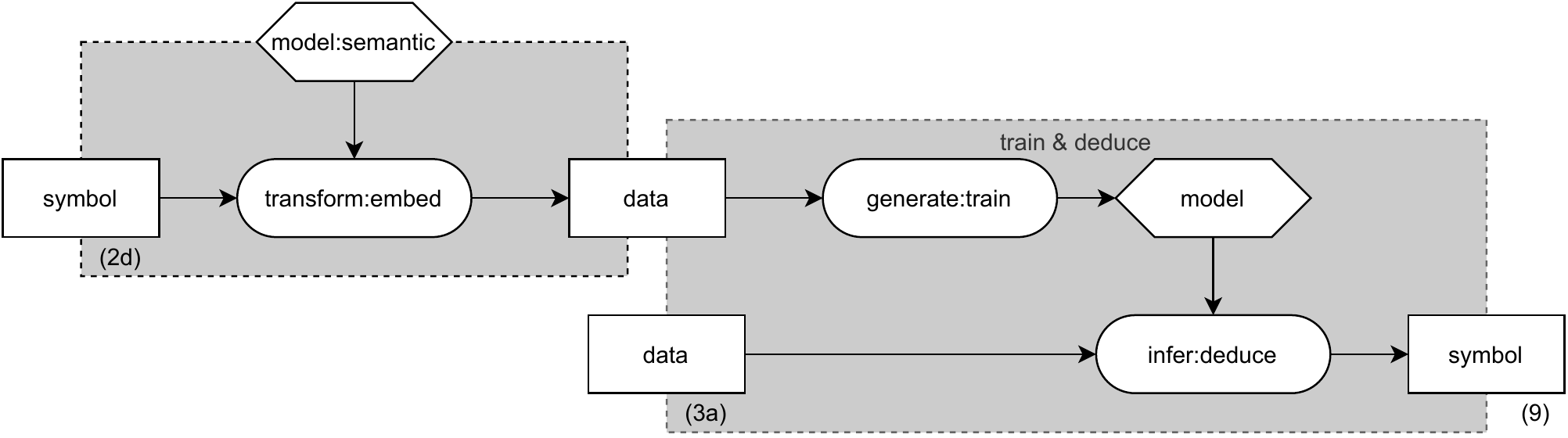}}

\subsection{Learning to reason}

\todo{hier een variant zonder "transform" toevoegen, zodat 1a ook 1b wordt.}
\todo{REPLACE THE LOWEST 2B by 2C}

\centerline{\includegraphics[scale=0.6]{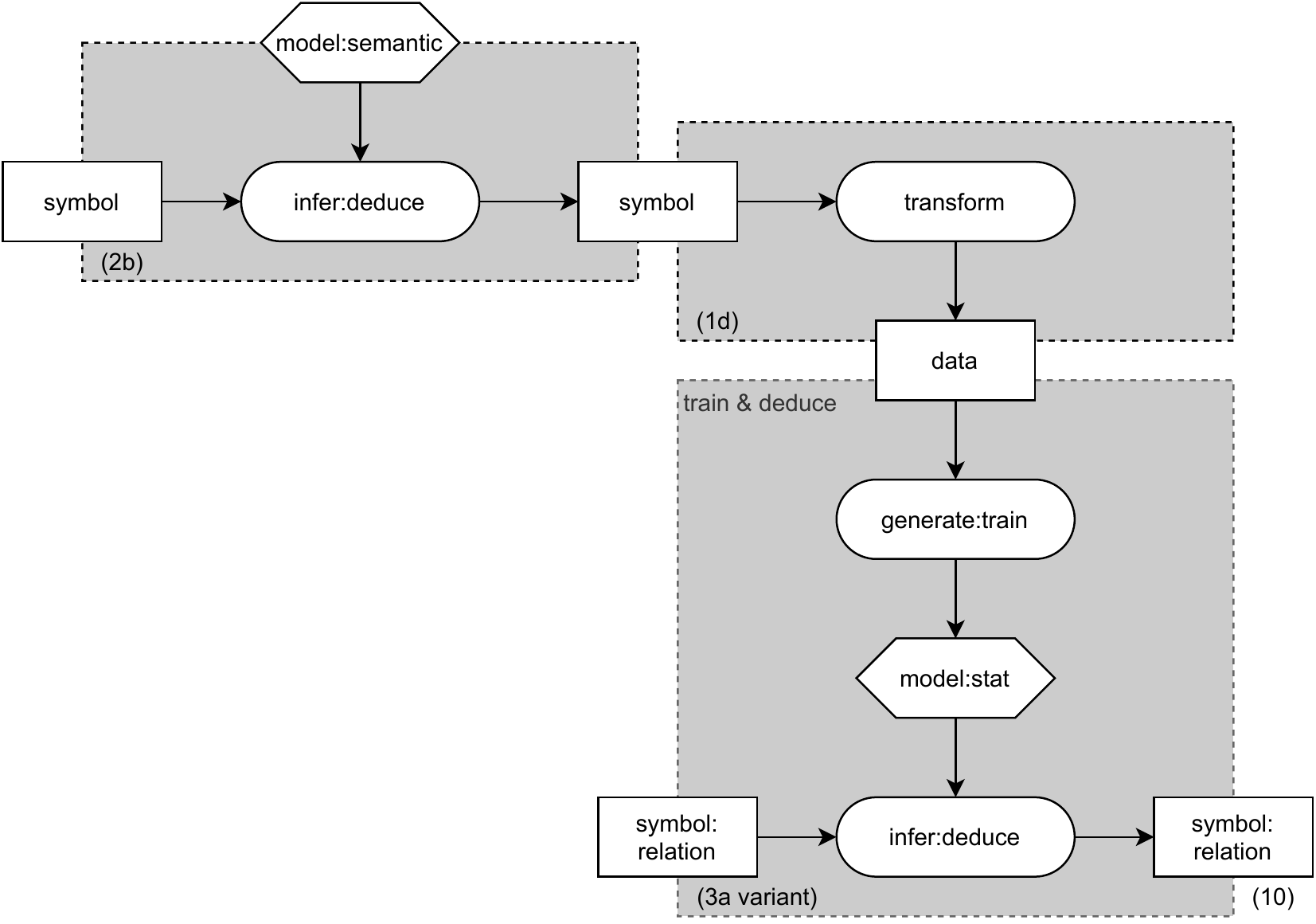}}

Whereas pattern (9) provides representation learning by embedding vector/tensor representations of logical structures in a neural network, there are also attempts to learn the reasoning process itself in neural networks. This is motivated by the ability of neural networks to provide higher scalability and better robustness when dealing with noise in the input data (incomplete, contradictory, erroneous). The focus of pattern (10) is on reasoning with first-order logic on knowledge graphs. This pattern learns specific reasoning tasks based on symbolic input tuples and the inferencing results from the symbolic reasoner. Pattern (10) is a combination of our basic patterns for symbolic reasoning (2b) and training to produce a statistical model (1a). 

This pattern for training a neural network to do logical reasoning captures a wide variety of approaches such as reasoning over RDF knowledge bases \cite{Hitzler-learning-to-reason}, Description Logic Reasoning \cite{hohenecker2017deep} and logic programming \cite{differentiable-proving}. 
Relational Tensor Networks (RTNs) \cite{hohenecker2017deep}
use a recurrent neural network to learn two kinds of predictions, namely the membership of individuals to classes and the existence of relations. In a somewhat different application, \cite{ebrahimi2018reasoning}
takes a set of normalized triples and a normalized query to learn classification of the entailment of the query from the statements in the current knowledge graph. 
Whereas current efforts focus on deductive reasoning in knowledge bases of FOL and DL (or fragments thereof), the pattern can in theory be applied to other inference tasks and mechanisms.

\subsection{Meta reasoning for control}

%11

\centerline{\includegraphics[scale=0.6]{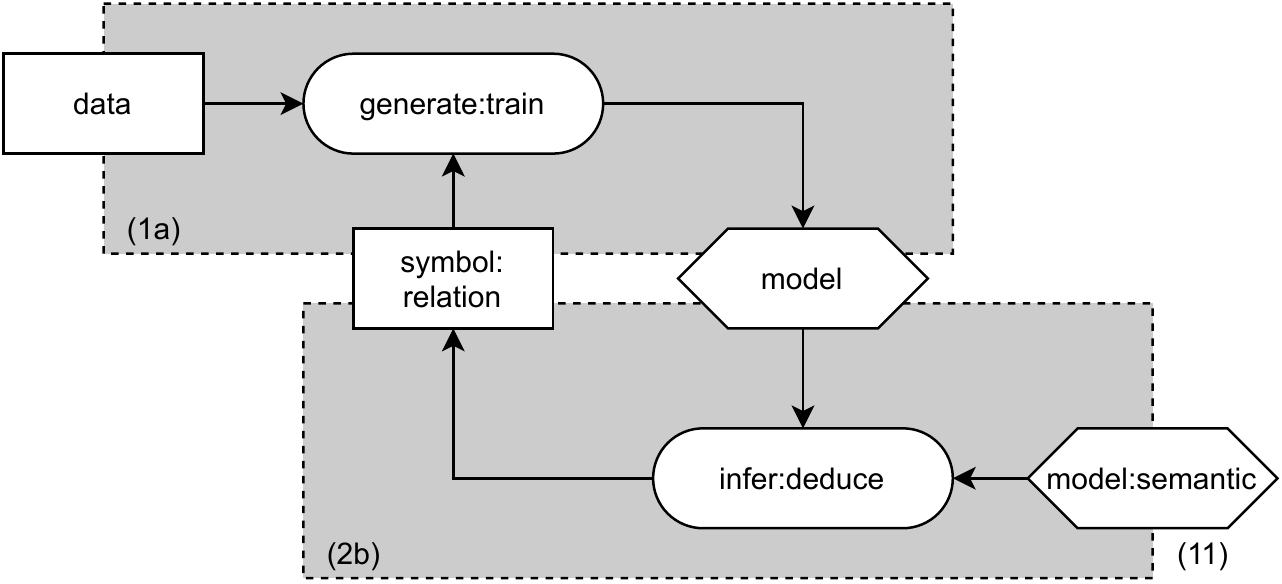}}

\todo{a better picture would have the 2b subsystem on top, and running from left-to-right}
\todo{(2b) has two models as input???}

There is a long-standing tradition in both AI \cite{meta-reasoning-survey} and in the field of cognitive architectures (e.g. \cite{PerlisBKM17}) to investigate so-called meta-reasoning systems, where one system reasons about (or:learns from) the behaviour of another system.

It is widely recognised that the configuration of machine learning systems is highly non-trivial, ranging from choosing an appropriate neural network architecture to setting a multitude of hyper-parameters for that architecture. The field of AutoML \cite{automl} aims to automate that configuration task. This is often done through applying machine learning techniques to this task (ie the system is learning the right hyper-parameter settings for the target learning system), but the configuration of the target system can also be done by capturing the knowledge of machine learning engineers. This is done in a system such as Alpine Meadow \cite{shangalpine}
and is captured in pattern (11): a knowledge based of ML configuration knowledge is used to deduce appropriate hyper-parameter settings for a learning system (sub-pattern \figcite{(2b)}), these parameters are then used to train a model (requiring a slightly modified version of sub-pattern \figcite{(1a)}), and the resulting performance of this model is inspected by the knowledge base which may give rise to adaptations of the hyper-parameters in the next iteration. 

Another class of systems which at first sight may seem very different, but which are an instantiation of the same pattern are so called "curriculum-guided learning" systems. Curriculum learning problem can be defined as the problem of finding the most efficient sequence of learning situations in these various tasks so as to maximize its learning speed \cite{Fournier:2018,Fang:2019}. In terms of pattern (11), the task of the (2b) subsystem is to feed the learner in subsystem (1a) its training instances in the optimal sequence.

% Annette: wat moet er met deze ref. ook al weer?
%% \cite{minervini2020learning}
\todo{Add the recent [PDF] Learning Reasoning Strategies in End-to-End Differentiable Proving
P Minervini, S Riedel, P Stenetorp, E Grefenstette - arXiv preprint arXiv, 2020, Attempts to render deep learning models interpretable, data-efficient, and robust have seen some success through hybridisation with rule-based systems, for example, in Neural Theorem Provers (NTPs). These neuro-symbolic models can }

\todo{redraw the figure of 4.8 to more closely resemble the layout of 4.4. One with with feedback from the learning process, and one is without} 

Notice that this pattern closely resembles pattern (7) (informed learning with prior knowledge). In that pattern, the symbolic component deduces prior knowledge as input for the training component once, but the resulting training model is not subsequently inspected to possibly adjust this input.
Whereas in pattern (11) a symbolic system is used to guide the learning behaviour of a subsymbolic system, the converse is also possible. In \cite{Loos}, a system is presented where a subsymbolic system learns the search strategies needed to guide a symbolic theorem prover. This line of work has a long history, dating back to the 1990's \cite{Suttner:1990}.
\todo{This would require an additional pattern 11a}

\section{Two Use-cases}
\label{sec:use-cases}

\centerline{\includegraphics[scale=0.6]{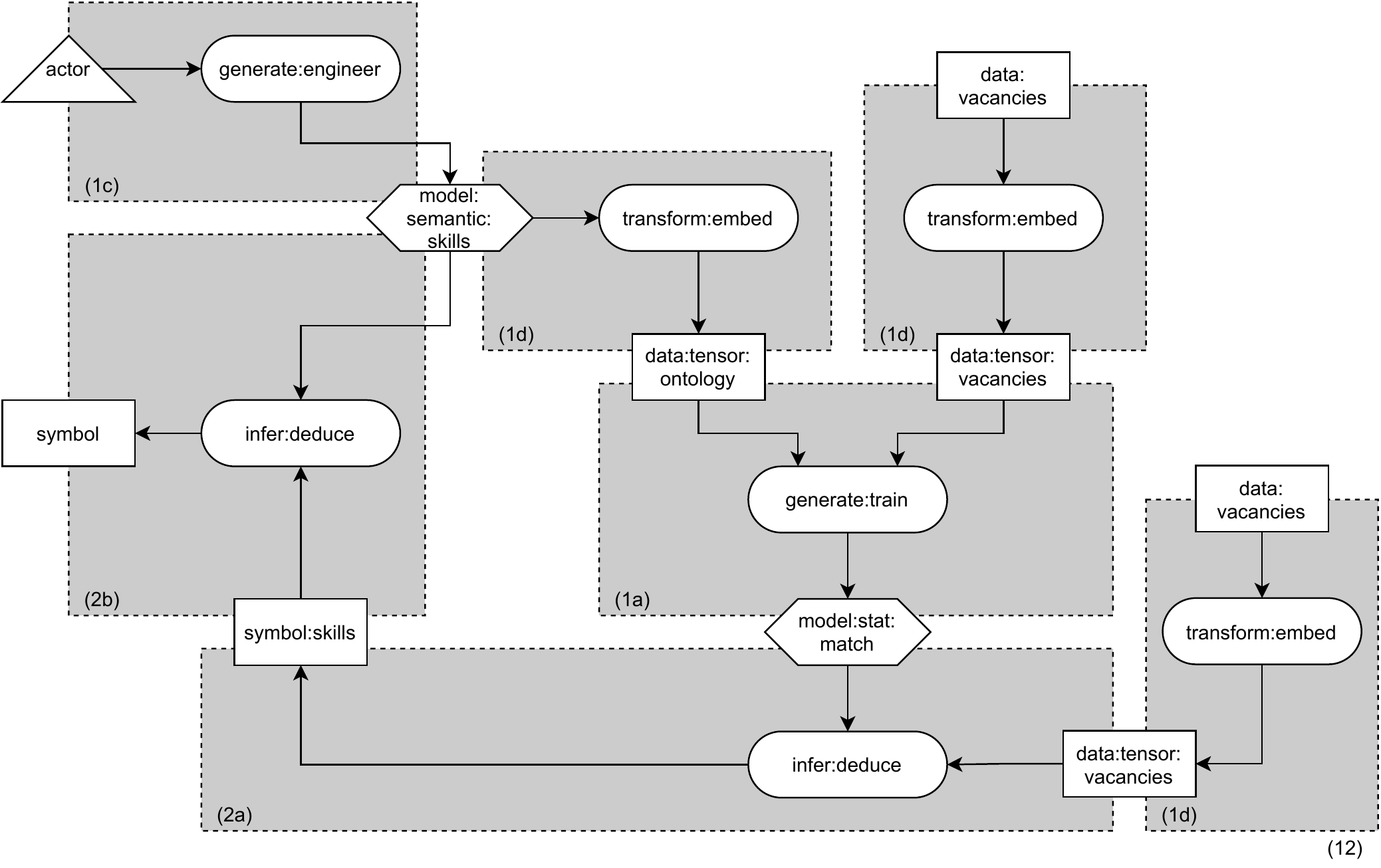}}

In this section, we describe the use of our boxology patterns in two real world use cases.

We have identified the need for a pattern-based system of systems approach towards design and evaluation of hybrid AI systems. Using our approach can increase the level of trustworthiness of such systems. Trust in AI systems emerges from a number of factors, such as transparency, reproducibility, predictability and explainability. A Hybrid AI system is not to be seen as a monolithic component, but communicating modules of such a hybrid system \cite{garcez2020neurosymbolic}. Insight into the individual modules and components and their relationships and dependencies is essential, in particular in a decentralised system. The specification and verification of each component and their interactions enable a system-wide validation of expected behaviour. The definition and use of best practices and design patterns supports the generation of trustworthy AI systems - either when building new systems or when understanding existing systems by dissection and reverse-engineering.

Our method allows for step-wise refinement of a system design by starting with a high level of abstraction and drilling down towards implementation details and reusable components by specifying more and more concrete choices, such as which models to use. Starting from generic patterns, an implementation can be derived and deployed, based on the experience and best practices in Hybrid AI.

\subsection{Skills Matching}

In the first use case, the goal is to create a piece of software that is able to match open vacancies or job descriptions with CVs of job seekers. In this specific use case skills, defined as the ability to perform a task, are used to do the matchmaking. In the first part of the project, a large architecture picture was created with a lot of boxes, arrows and terms. The distinction between processes and data was not very clear and the type of model was also not explicitly defined. 

With use of the boxology, the architecture picture is better readable (see figure (12)). This figure made it possible to on the one hand talk about the bigger boxes (elementary patterns) and on the other hand go into more detail about the specific implementation of for example a model box and the specific input and output type needed. The figure makes it also possible to think about the future of the project in terms of patterns that have to be added, substituted or removed.

To go into more detail for this specific use case: in the training phase, vacancies are transformed to the data type tensor using an (word2vec) embedding. A Skills ontology, which is a semantic model, engineered by a human {(pattern 1c)}, is also transformed to a tensor using an embedding {(variant of pattern 1d)}. Both the vacancy and the skill ontology are used to train a neural network {(pattern 1a)}, a statistical model. This model learns which sentences or parts of sentences of a vacancy text contain a skill and match to which specific skill in the ontology. When this model is applied to a new vacancy, transformed using an embedding to a tensor {(pattern 1d)}, it predicts the most probable skill through deduction {(pattern 2a)}. An additional step is to use the ontology to deduce {(pattern 2b)} the label skill to a more understandable skill, for example a skill with a description. 

\subsection{Robot in Action}

\centerline{\includegraphics[scale=0.6]{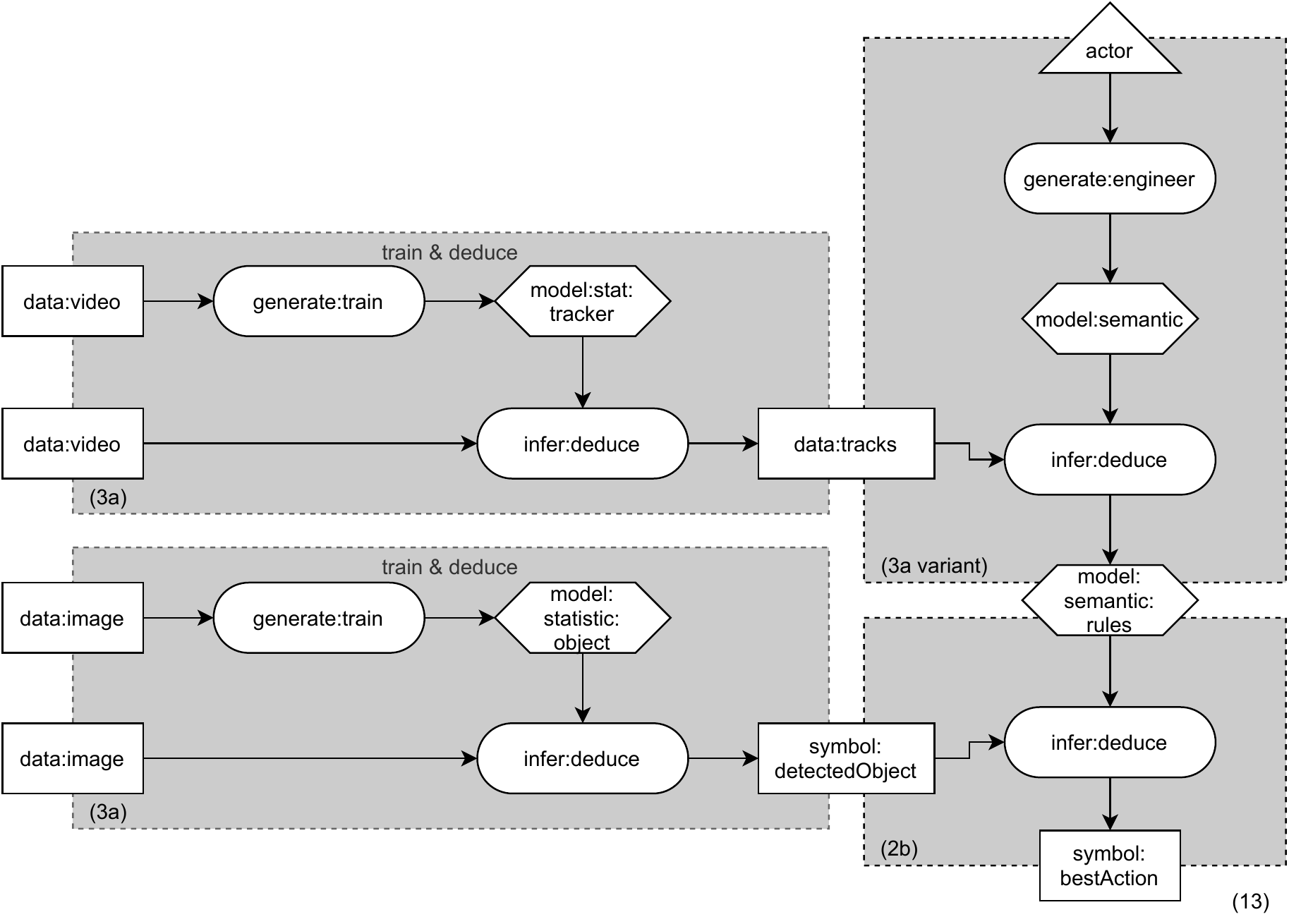}}

In the second use case, the goal is to get a robot to choose the best action to perform. The robot has access to a camera. In a training phase, an object detector and a tracker are trained using statistical models, whereas an ontology is handcrafted by an expert actor. The object detector is a neural network trained using images of the environment {(pattern 1a)}. The tracker uses the video stream of the camera and uses a different type of neural network {(pattern 1a)}. A semantic model in the form of an ontology about the world is engineered by a human {(pattern 1c)}. At each timestep, the robot obtains new information from the camera. The image is used to predict which objects are visible in the environment {(pattern 2a)}. The tracker is used to track objects through time {(pattern 2a)}. The tracks and the world model are used to induce semantic rules {(pattern 2c)}, such as reasoning what will happen next. Then these rules and the detected object(s) are used to predict what the best action is {(pattern 2b)}.

In this use case, the boxology helped to see the similarity between the object detector and the tracker and to determine how the detailed process flow should be.

\section{Related Work}

In \cite{von2019informed} a large collection of over 100 different systems for ``informed machine learning'' is presented.  The survey  paper provides a broad overview of how many different learning algorithms can be enriched with prior knowledge. Their figure 2 provides a taxonomy of "informed machine learning" across three dimensions: which source of knowledge is integrated (e.g. expert knowledge or common sense world knowledge), how that knowledge is represented (e.g. as knowledge graphs, logic rules, algebraic equations and 5 other types of representation) and where this knowledge is integrated in the machine learning pipeline (in the training data, the hypothesis set, the learning algorithm or the final hypothesis). The second and third of these dimensions are used to categorise well over 100 different published systems into an 8x4 grid. All of these systems are captured in one of our design patterns (pattern 7), so while our proposal covers a much broader set of hybrid systems, the result of \cite{von2019informed} is a very detailed analysis of one of our patterns.

In his invited address to the AAAI 2020 conference, Henry Kautz introduced a taxonomy for neural-symbolic systems\footnote{\url{https://www.cs.rochester.edu/u/kautz/talks/index.html}} The proposed taxonomy consists of a flat set of 6 types. We will briefly summarise (our understanding of) these informally characterised types (partly based on the explanation in \cite{garcez2020neurosymbolic}), and we will show how Kautz' types relate to the design patterns that we will propose in this paper. Type 1 systems (informally written by Kautz as ``symbolic Neuro symbolic'') are learning systems that have symbols as input and output. This directly corresponds to our elementary pattern 3b. Type 2 systems (informal notation: ``Symbolic[Neuro]'') are symbolic reasoning systems that are guided by a learned search strategy. These directly correspond to a variation of our pattern (11). Type 3 systems (informal notation ``Neuro;Symbolic'') consist of a sequence of a neural learning system that performs abstraction from data to symbols, followed by a symbolic system. This corresponds to our patterns (6a) and (6b), showing that in this case we make a more fine-grained distinction. Type 4 systems (informal notation ``Neuro:Symbolic $\rightarrow$ Neuro'') use symbolic I/O training pairs to teach a neural system. These correspond partly to our elementary pattern (3b) (for example: inductive logic programming), partly to our pattern (8) (eg link prediction) and partly to pattern (10) (eg learning to reason), again showing that we propose a much more fine-grained distinction. Type 5 systems (informal notation ``Neuro'') use symbolic rules that inform neural learning. These correspond to our pattern 7. Finally, Type 6 systems (informally ``Neuro[Symbolic]'') remains somewhat unclear (as also acknowledged in \cite{garcez2020neurosymbolic}), and we refrain from interpreting this type.\\

\begin{center}
\begin{tabular}{|c|c|c|c|c|c|c|}
\hline
Kautz types: & Type 1 & Type 2 & Type 3 & Type 4 & Type 5 & Type 6 \\
\hline
Our Patterns: & 3b & 11 & 6a,6b & 3b, 8, 10 & 7 & - \\
\hline
\end{tabular}
\end{center}

\noindent
The above table shows that there are substantial differences between our proposed design patterns and the system types from Kautz. Kautz' taxonomy has similar goals to ours, namely to identify different interaction patterns between neural and symbolic components in a modular hybrid architecture, but our proposal goes beyond Kautz' proposal because (a) Kautz proposes a taxonomy of systems without describing the internal architectures of the types of systems in his taxonomy, and (b) we make more fine-grained distinctions than Kautz, refining his 6 categories into distinctive subtypes, each with their own internal modular architecture (= design pattern).

In \cite{DeRaedtIJCAI2020} the authors survey hybrid (``neural-symbolic'') systems along eight different dimensions. We briefly describe each of these, and discuss their relationship to the distinction made in our own work. 
\begin{description}
    \item[\textit{Directed vs undirected graphical models}.] 
This is a finer grained distinction about representations than we make. In our patterns, these are both captured by the same ``semantic model'' component 
\item[\textit{Model-based vs. proof-based inference}] (better known as model-theoretic vs. proof-theoretic inference). Note this use of ``model-based'' is using ``model'' as the term is used by logicians, which is unlike how Darwiche uses the term model-based, using the term model as used by machine learning, showing the highly ambiguous use of the term ``model'' in Computer Science in general and in AI in particular.
Again, this is a finer grained distinction than we make, and again both of these forms of inference are captured in our single KR component.
\item[\textit{Logic vs. Neural}.]
This corresponds to our distinction between ML and KR components
\item[Boole\textit{an vs. probabilistic semantics.}]
Similar to above, both of these would be captured by the KR component without making the distinction
\item[\textit{Structure vs. parameter learning}.]
This is captured in our notation by ML components that have either a statistical or a semantic model as their result
\item[\textit{Symbols vs. Sub-symbols}.]
This corresponds to our distinction between symbols and data. Unfortunately, and similar to our work, De Raedt et al. do not give a precise distinction between the two categories. 
\item[\textit{Type of Logic}.]
This is another finer distinction then we make, these different types of logic are all captured in terms of our KR component
\end{description}
Summarising, on the one hand, De Raedt et al. make a number of finer distinctions than our boxology, mostly in terms of the details inside our components (different variations of KR components, different variations of models), while on the other hand de Raedt et al. do not discuss how these components should be configured into larger systems in order to achieve a particular functionality, which is the goal of our boxology. 
Whereas our boxology is a refinement of the 6 types proposed by Kautz (both aiming to describe modular architectures of interacting components), the work by de Raedt et al. is a refinement of some of our components, and could be combined with our work in a future version.  The same is true for  \cite{von2019informed}.

\section{Conclusion and Future Work}
\label{sec:future-work}

In our paper, we have presented a visual language (boxology) to represent learning and reasoning systems. The taxonomical vocabulary and a collection of patterns expressed in this language aim to foster a better understanding of Hybrid AI systems and support communication between AI communities. A practical application in two use cases demonstrates that we can use the boxology to create a communicable blueprint of rather complex Hybrid AI systems that integrate various AI technologies.

The work presented here provides ample opportunities for additional features and uses. We expect to apply the taxonomy and visual language in many more use cases and is likely to evolve further as a result. New examples of AI systems will contribute to extending and improving the taxonomy, which in turn allows us to cover more use cases. Using this approach, an increasingly more mature visual language will evolve.

As a first extension to the current boxology, the concept of actors can be defined, along with the corresponding interaction processes and models. Actors are necessary for modelling interactions among autonomous entities, such as software agents or robots, whether they are physically or logically distributed. They also allow for specifying systems with humans in the loop and human-machine interaction in general. Use cases for actors include federated learning and reasoning, multi-robot coordination or hybrid human-agent teams. 
In \cite{Witschel2021} the authors propose an extension of our boxology \cite{VanHarmelen2019Boxology} with two abstract patterns for humans-in-the-loop systems, namely where the human agent either performs the role of a feedback-provider or a feedback-consumer. 
\begin{comment}
Furthermore, dealing with confidential data and local knowledge can be used for resource management (e.g., using negotiation) in IoT systems, ecosystems of apps and services, and for modelling (normative) behaviour in social simulations.
\todo{Mvb: what does the last sentence refer to? Is this another idea for extending?}
\end{comment}

Future work also includes developing the boxology from a means of representing system functionality towards an architectural tool-set of reusable components for design, implementation and deployment of hybrid AI systems. A more coherent methodology for complex AI systems based on the boxology allows these systems to be easier to understand in terms of functionality. This in turn provides a basis for more explainable and trustworthy AI systems design. 
An interesting topic to pursue in this respect is the creation and development of a generative grammar and logic calculus for composing and verifying patterns. This would facilitate the above-mentioned goals of and allow for formal verification at the component, pattern and system levels.

When using a coherent methodology for complex Hybrid AI system design, it is expected that such a design becomes easier to understand and maintain. In addition, hybrid AI systems will become better explainable, responsible, reliable and predictable. It is our aim to develop such systems as being trustworthy by design. This could provide a framework for system quality control by evaluation and certification.

Finally, the methodology needs to be further documented using guidelines for specifying increasingly concrete implementations of the concepts. 

\section*{Acknowledgement}

This research was partially funded by the Hybrid Intelligence Center, a 10-year programme funded
by the Dutch Ministry of Education, Culture and Science through the Netherlands Organisation for
Scientific Research NWO, \url{https://hybrid-intelligence-centre.nl}.

\bibliographystyle{abbrv} 
\bibliography{references}

%%%%%%%%%%%%%%%%%%%%%%%%%%%%%%%%%%%%%%%%%%%%%%%%%%%%%%%%%%%%%%%%%%%%%%%%%%%%%%%%%%%%

\section*{Appendix: Taxonomy}

\begin{longtable}{|p{0.25\textwidth}|p{0.75\textwidth}|}
\hline
\textbf{Concept} & \textbf{Definition}   \\
\hline   
\textbf{Instance}                     & An example or single occurrence of something. (Collins)                                 \\
\hspace*{0.5cm}\textbf{Data}                         & Factual information (such as measurements or statistics) used as a basis for reasoning, discussion, or calculation. (MW)                                                     \\
\hspace*{1.0cm}Number                                & A numerical quantity that is assigned or is determined by calculation or measurement. (MW)                                                                                  \\
\hspace*{1.0cm}Text                                  & Words and form of a written or printed work. Something (such as a story or movie) considered as an object to be examined, explicated, or deconstructed. (MW)                           \\
\hspace*{1.0cm}Tensor                                & A set of components, functions of the coordinates of any point in space, that transform linearly between coordinate systems. For three-dimensional space there are 3 r components, where r is the rank. A tensor of zero rank is a scalar, of rank one, a vector. (Collins) A picture (bitmap) can be represented as a tensor of rank 2.  \\
\hspace*{1.0cm}Stream                                & A stream of things is a large number of them occurring one after another. (Collins) Digital data (such as audio or video material) that is continuously delivered one packet at a time and is usually intended for immediate processing or playback. (MW)                                                                                   \\
\hspace*{0.5cm}\textbf{Symbol}                       & Something that stands for or suggests something else by reason of relationship, association, convention, or accidental resemblance. An arbitrary or conventional sign used in writing or printing relating to a particular field to represent operations, quantities, elements, relations, or qualities (MW)  \\
\hspace*{1.0cm}Label                                 & A descriptive or identifying word or phrase (MW)                                             \\
\hspace*{1.0cm}                              & An aspect or quality (such as resemblance) that connects two or more things or parts as being or belonging or working together or as being of the same kind. (MW)                                                 \\
\hspace*{1.0cm}Trace                                 & A sign or evidence of some past thing. (MW)                                                   \\
\hline
\textbf{Model}                        & A system of postulates, data, and inferences presented as a mathematical description of an entity or state of affairs. (MW)                                                                                                                                                                                                  \\
\hspace*{0.5cm}\textbf{Statistical Model}            & A statistical model is usually specified as a mathematical relationship between one or more random variables and other non-random variables. (MW)                                                                                                                                                                            \\
\hspace*{0.5cm}\textbf{Semantic Model}                  & A conceptual model represents 'concepts' (entities) and relationships between them. (MW) Semantic technologies formally represent the meaning involved in information. For example, ontology can describe concepts, relationships between things, and categories of things. (MW)  \\
\hline
\textbf{Processing}                   & A series of actions or operations conducing to an end. (MW)                                                                                                                                                                                                                                                                  \\
\hspace*{0.5cm}\textbf{Generation}                   & A process of coming or bringing into being. (MW) Defining or originating (something, such as a mathematical or linguistic set or structure) by the application of one or more rules or operations. (MW)                                                                           \\
\hspace*{1.0cm}Training                              & The process of learning the skills that you need for a particular job or activity. (Collins) In particular: model building, the process of preparing a machine learning model to be useful by feeding it data from which it can learn, ie. detect statistical patterns.         \\
\hspace*{1.0cm}Engineering                         & The application of science and mathematics by which the properties of matter and the sources of energy in nature are made useful to people. The design and manufacture of complex products. (MW) In particular, software engineering: the methodical design, implementation, and maintenance of models and components of software architectures.                   \\
\hspace*{0.5cm}\textbf{Transformation}              & The operation of changing (as by rotation or mapping) one configuration or expression into another in accordance with a mathematical rule. (MW)                                                                                                                                                                              \\
\hspace*{0.5cm}\textbf{Inference}                    & A conclusion or opinion that is formed because of known facts or evidence. (A) The act of passing from one proposition, statement, or judgment considered as true to another whose truth is believed to follow from that of the former. (B) The act of passing from statistical sample data to generalisations, usually with calculated degrees of certainty. (MW)\\
\hspace*{1.0cm}\textbf{Induction}  & Inference of a generalised conclusion from particular instances. (MW)\\
\hspace*{1.0cm}\textbf{Deduction}  & The deriving of a conclusion by reasoning.
Inference in which the conclusion about particulars follows necessarily from general or universal premises. (MW)\\

\hspace*{1.5cm}\textbf{Classification}   & systematic arrangement in groups or categories according to established criteria. (MW)\\

\hspace*{1.5cm}\textbf{Prediction}   &To calculate (some future event or condition) usually as a result of study and analysis of available pertinent data. (MW)\\

\hline
\textbf{Actor}   &Autonomous entity acting proactively to initiate processes, based on intentions and goals. Interaction among actors leads to emergent system behaviour.\\
\hspace*{0.5cm}\textbf{Human}   &Human beings interacting with AI system components and other actors.\\
\hspace*{0.5cm}\textbf{Agent}   &Active software components that are proactive, rather than merely reactive. Agents can be based on many internal reasoning mechanisms.\\
\hspace*{0.5cm}\textbf{Robot}   &Physically embedded agents that can sense their environment and act in it.\\
\hline
\caption{Hybrid AI taxonomy}
\label{tab:taxonomy}
\end{longtable}

\end{document}